\RequirePackage[svgnames]{xcolor}
\documentclass[twocolumn]{mystyle}

\usepackage[all]{hypcap}
\usepackage[svgnames]{xcolor}

\usepackage[numbers]{natbib}
\usepackage{hyperref}

\usepackage{algorithm}
\usepackage{algorithmicx}
\usepackage{algpseudocode}
\usepackage{microtype}
\usepackage{graphicx}
\expandafter\def\csname ver@subfig.sty\endcsname{}
\usepackage{booktabs} %
\usepackage{float}
\usepackage{bigstrut}

\usepackage{amsmath}
\usepackage{amssymb}
\usepackage{mathtools}
\usepackage{amsthm}
\usepackage{mathrsfs}
\usepackage{nicefrac}
\usepackage{dsfont}
\usepackage{enumitem}
\usepackage{subcaption}
\usepackage{graphicx,subfig}
\usepackage{cleveref}
\usepackage{bxcoloremoji}

\usepackage{float}

\usepackage[utf8]{inputenc} %
\usepackage[T1]{fontenc}    %
\usepackage{hyperref}       %
\usepackage{url}            %
\usepackage{booktabs}       %
\usepackage{amsfonts}       %
\usepackage{nicefrac}       %
\usepackage{microtype}      %
\usepackage{graphicx}
\usepackage{subcaption} 
\usepackage{fdsymbol}
\usepackage{wrapfig}
\usepackage{lipsum}
\usepackage{enumitem}
\usepackage{stackengine}
\usepackage[font=small,labelfont=bf]{caption}
\usepackage{color}
\usepackage{adjustbox}

\usepackage{rotating}
\usepackage{makecell}
\usepackage{xspace}
\usepackage{multirow}
\usepackage{listings}
\definecolor{ForestGreen}{RGB}{34,139,34}

\renewcommand{\paragraph}[1]{\medskip\noindent\textbf{#1.~}}

\newcommand{\github}{\raisebox{-1.5pt}{\includegraphics[height=1.05em]{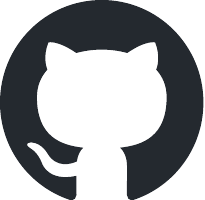}}}

\newcommand{\modelname}{\textbf{\textsc{DocMemo}}}
\usepackage{amsmath}
\usepackage{lineno}

\title{\modelname{}: Dynamic Evidence Discovery via Probabilistic Memory-Guided Retrieval for Multi-Modal Document Understanding}
\runningtitle{\modelname{}: Dynamic Evidence Discovery via Probabilistic Memory-Guided Retrieval for Multi-Modal Document Understanding}

\author[1]{Hanshu Yao}
\author[1]{\mbox{Jianfeng Zhong}}
\author[2]{\mbox{Niu Lian}}
\author[1,\dagger]{\mbox{Jinpeng Wang}}
\affil[1]{\mbox{Harbin Institute of Technology, Shenzhen}}
\affil[2]{\mbox{Tsinghua Shenzhen International Graduate School, Tsinghua University}}

\authornote[\dagger]{\mbox{Corresponding Authors}}

\begin{document}
\begin{abstract}
Long-document understanding requires locating sparse and heterogeneous evidence across hundreds of pages, yet existing systems remain limited by static retrieval and fragile cross-round memory. Mainstream single-round methods commit to a fixed top-$k$ page set at the outset and struggle to recover from early retrieval errors; recent iterative approaches allow multi-round evidence acquisition, but they do not investigate the propagation mechanism of cross-round states, making it difficult to track the dynamic changes in page relevance. To address these limitations, we propose DocMemo, a memory-guided framework that formulates long-document reasoning as dynamic evidence exploration. DocMemo maintains a tri-level retrieval state consisting of Document Schema Memory, Page Belief Memory, and Question Episodic Memory, which respectively capture structural priors, dynamic relevance estimation, and query-specific reasoning trajectories. During reasoning, \modelname{} continuously refines cross-round page selection through Bayesian page belief updating with Thompson sampling, spatial proximity propagation, and structure-aware adaptive-granularity evidence access, while supplementing page-level evidence with fine-grained visual regions. Experiments on 3 benchmarks show that \modelname{} achieves state-of-the-art performance and validate the efficacy of structured memory and dynamic page belief updating. 

\vspace{2mm}
\vspace{5mm}

\coloremojicode{1F4C5} \textbf{Date}: August 7, 2026

\github{} \textbf{Code Repository}: \href{https://github.com/Harrygof/DocMemo}{https://github.com/Harrygof/DocMemo}

\coloremojicode{1F4E7} \textbf{Contact}: \href{mailto:harrygof55@gmail.com}{harrygof55@gmail.com} (Hanshu Yao) \href{mailto:wangjp26@gmail.com}{wangjp26@gmail.com} (Jinpeng Wang)

\end{abstract}

\maketitle
\section{Introduction}

\label{sec: introduction}
\begin{figure}[!t]
    \centering
    \includegraphics[width=\linewidth]{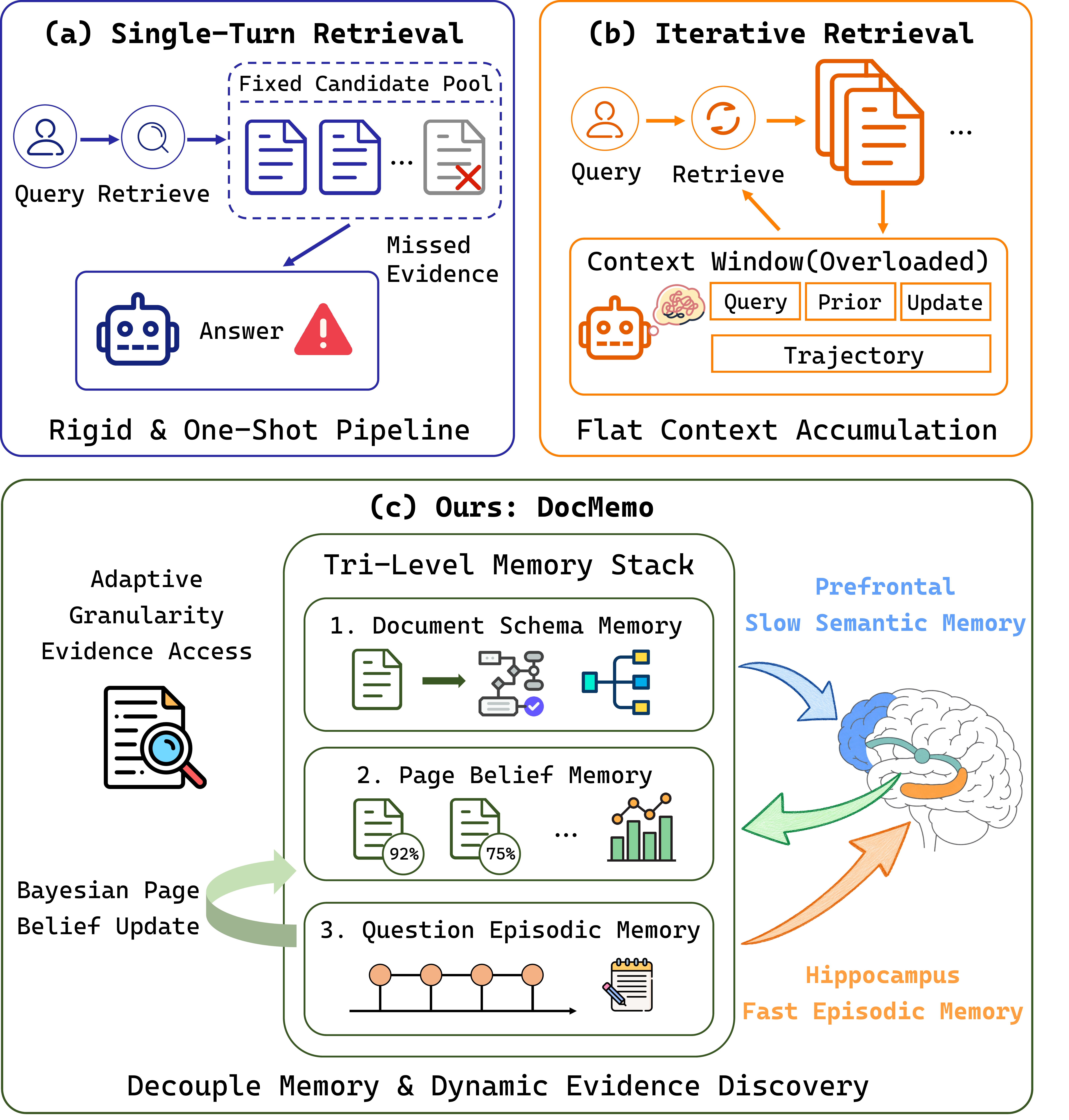}
    
    \caption{
Overview and motivation of \modelname{}. 
(a) Single-turn static retrieval commits to a fixed candidate page pool before reasoning; once key evidence is missed, the model cannot revise its search trajectory. 
(b) Iterative retrieval allows multiple search rounds, yet it lacks a cross-round state propagation mechanism.
(c) \modelname{} introduces a tri-level memory structure. Reasoning feedback is written back through Bayesian page belief updating and spatial proximity propagation, enabling memory-guided dynamic evidence discovery and adaptive-granularity evidence access.
}
    \label{fig:introduction}
\end{figure}
Long-document visual question answering (DocVQA) requires models to locate dispersed and heterogeneous evidence across documents spanning tens or even hundreds of pages. Such evidence may appear in text, tables, figures, and visually structured layouts. Since document length far exceeds the context budget of existing models, the central challenge of long-document DocVQA is not passive reading of long inputs, but dynamically locating, updating, and integrating relevant pages under a limited evidence budget~\cite{cho2024m3docrag}.

However, existing systems still struggle with such stateful exploration. Figure~\ref{fig:introduction} contrasts existing retrieval paradigms with the core design of \modelname{}. Single-turn static retrieval methods typically fix a candidate page set before reasoning; once key evidence is missed in the initial retrieval stage, subsequent reasoning has little opportunity to recover it~\cite{lewis2020retrieval, karpukhin2020dense, faysse2025colpali}. Recent methods improve candidate-page recall through richer multimodal indexing, graph-based document organization, or cross-page relation modeling, but their page selection is usually completed before reasoning and thus cannot be easily adjusted based on intermediate reasoning signals~\cite{zhu2025doclens, tanaka2025vdocrag, yu2025visrag, cho2024m3docrag}. Iterative methods~\cite{jain2025simpledoc} allow retrieval over multiple rounds, but they lack cross-round state propagation mechanisms. Cross-round information is mainly maintained through context reconstruction, making the process closer to repeated independent retrieval~\cite{asai2024self, jiang2023active, trivedi2023interleaving}. As a result, although retrieval is formally extended to multiple rounds, it still lacks a structured exploration memory that can be continuously updated.

To bridge this gap, we propose \modelname{}, a memory-guided long-document reasoning framework that explicitly models persistent document knowledge, dynamic page relevance, and query-specific exploration history. This design is motivated by the complementary learning principle of separating stable and rapidly changing information.~\cite{mcclelland1995there}. \modelname{} introduces dynamic probabilistic state modeling into long-document reasoning and constructs a tri-level document memory: \textit{Document Schema Memory} models document structural priors, \textit{Page Belief Memory} maintains page-level relevance confidence that evolves with cross-round feedback, and \textit{Question Episodic Memory} records intermediate findings and search trajectories under the current query~\cite{dong2025towards}. Through this explicit memory modeling, \modelname{} transforms iterative retrieval from context accumulation into structured evidence exploration jointly driven by persistent document knowledge, dynamic page states, and query-local experience~\cite{xu2026mem, yan2025memory}.

To enable dynamic probabilistic state modeling, \modelname{} introduces Bayesian page belief updating, which formulates page selection as a confidence updating process that continuously evolves with reasoning feedback. Combined with Thompson sampling~\cite{thompson1933likelihood}, the system balances the exploitation of high-confidence pages with the exploration of uncertain candidates. We also design a spatial proximity propagation mechanism motivated by the locality of evidence distribution in long documents~\cite{sweller1988cognitive}, which leverages the structural regularity that evidence in long documents often clusters within local page ranges. In addition, \modelname{}{} adopts adaptive-granularity evidence access, supplementing fine-grained visual evidence to enhance complex layout reading.~\cite{wang2026agenticocr, li2026regionrag}.

We evaluate \modelname{} on MMLongBench-Doc, LongDocURL, and PaperTab~\cite{ma2024mmlongbench, deng2025longdocurl, hui2024uda}. Experimental results show that \modelname{} consistently improves over strong static retrieval and iterative baselines. Ablation studies further verify the effectiveness of tri-level memory and Bayesian belief updating, while analyzing the impact of fine-grained evidence augmentation. Overall, these results demonstrate that explicitly modeling and continuously updating document exploration states is key to improving long-document reasoning.

Our contributions are summarized as follows:
\begin{itemize}[leftmargin=*]
    \item We propose \modelname{}, a memory-guided framework for multi-modal document understanding, which organizes document exploration with a tri-level memory separating document priors, page beliefs, and query-specific reasoning traces.
    \item We introduce Bayesian page belief updating with Thompson sampling and spatial contiguity propagation, enabling uncertainty-aware and locality-aware dynamic retrieval across rounds.
    \item Experiments on three long-document DocVQA benchmarks show that \modelname{} consistently outperforms strong static and iterative baselines, with ablations validating the effectiveness of its core components.
\end{itemize}

\section{Related Work}
\label{sec:related_work}

\subsection{Long Document Visual Understanding}

Recent advances in long-document visual understanding have largely followed a retrieval-augmented reasoning paradigm~\cite{lewis2020retrieval, gao2023retrieval}. Existing methods can be broadly divided into two categories. Static retrieval-augmented approaches first select a fixed set of candidate evidence and then rely on multimodal models for answer generation; subsequent work further incorporates visual retrievers, structured retrieval, or fine-grained localization to improve evidence acquisition in complex documents ~\cite{cho2024m3docrag, wu2025molorag, yu2025visrag, wang2026agenticocr}. However, their retrieval decisions are typically fixed before reasoning begins, making it difficult to revise subsequent search based on intermediate clues. Iterative evidence retrieval offers an alternative direction. SimpleDoc, for instance, allows an agent to continue retrieving when current evidence is insufficient. However, it lacks a mechanism for transferring states across rounds. Cross-round information is primarily preserved through context reconstruction, making the process closer to repeatedly performing independent retrieval.~\cite{jain2025simpledoc, asai2024self, liu2024ra}.

In contrast, \modelname{} formulates long-document evidence acquisition as a structured memory-guided dynamic retrieval process. Its tri-level document memory explicitly organizes structural priors, evolving page beliefs, and query-specific reasoning traces, enabling retrieval states to be continuously refined across multiple exploration rounds.
\subsection{Memory for Agents}
Memory mechanisms have been extensively studied in LLM-based agents, with prior work exploring hierarchical storage and context management, long-term memory updating and forgetting, and associative memory organization~\cite{packer2023memgpt, xu2026mem, yan2025memory}. More recently, multimodal agent memory has gained increasing attention for supporting long-horizon perception and complex reasoning~\cite{he2024ma, long2025seeing, huang2025licomemory, lian2026verbatim}. However, existing frameworks are largely designed for open-ended interaction or temporally unfolding multimodal experience, with an emphasis on storing, compressing, and reusing historical information. They remain less effective at organizing complex multimodal content in which textual semantics, visual elements, and structural cues are tightly intertwined. This leaves the construction of structured memory capable of jointly organizing heterogeneous multimodal signals an important open problem for agent memory research~\cite{xie2024large, zhang2024document}.
\section{Method}
\label{sec:method}

\begin{figure*}[!t]
    \centering
    \includegraphics[width=\textwidth]{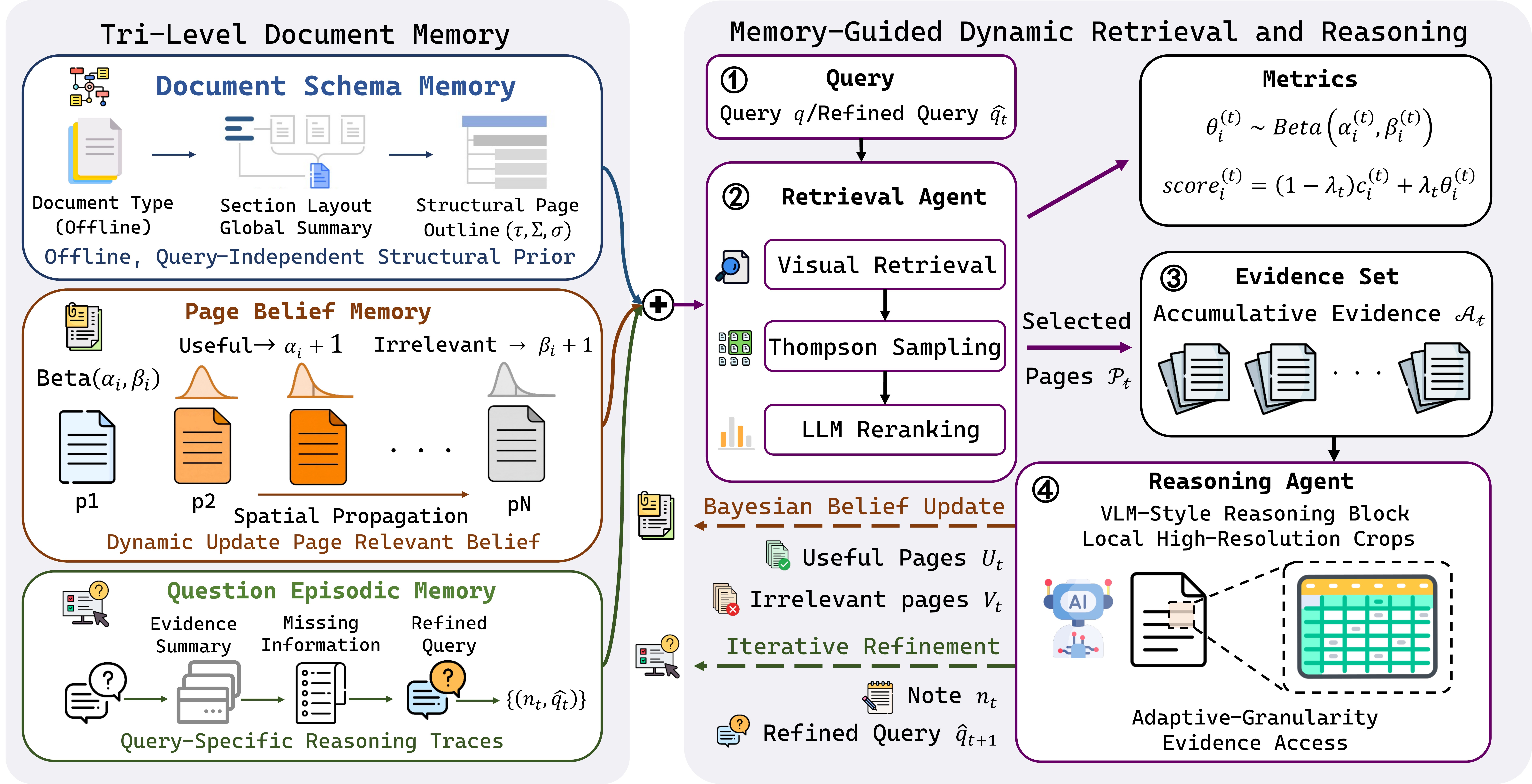}
    
    \caption{Overview of \modelname{}: (left) tri-level document memory maintains document schema, page belief, and question episodic traces; (right) memory-guided dynamic retrieval uses Thompson sampling and Bayesian belief updating to iteratively select evidence pages, incorporate reasoning feedback, and access fine-grained local regions for answer generation.}
    \label{fig:pipeline}
\end{figure*}

We propose \modelname{}, a memory-guided iterative reasoning framework for long-document visual question answering. \modelname{} organizes document exploration into three complementary memory states: Document Schema Memory provides reusable structural priors, Page Belief Memory maintains dynamically updated page-level relevance confidence, and Question Episodic Memory records evidence discoveries and search trajectories under the current query. Building upon this structured memory, \modelname{} progressively discovers document evidence through retrieval--reasoning interaction. Specifically, \modelname{} leverages Thompson sampling to dynamically select candidate pages under uncertainty and employs adaptive-granularity evidence access for local information understanding. The reasoning feedback is then incorporated through Bayesian page belief updating to refine page confidence and guide subsequent retrieval. Overall, \modelname{} forms a retrieval--reasoning--memory update process for continual evidence discovery in long documents.

\subsection{Tri-Level Document Memory}

Given a document \(D=\{p_i\}_{i=1}^{N}\) with \(N\) pages and a query \(q\), \modelname{} generates an answer through at most \(T\) retrieval--reasoning rounds. The system first pre-computes page-level visual embeddings and page summaries offline, and constructs a query-independent document structure representation. During online inference, the tri-level document memory jointly maintains the cross-round exploration state. Following the complementary learning principle of separating persistent and dynamic information, we represent the memory state at round \(t\) as
\[
\mathcal{M}^{(t)}
=
\left(
\mathcal{M}_{\text{schema}},
\mathcal{M}_{\text{belief}}^{(t)},
\mathcal{M}_{\text{epi}}^{(t)}
\right).
\]
Here, \(\mathcal{M}_{\text{schema}}\) models query-independent document structural priors, \(\mathcal{M}_{\text{belief}}^{(t)}\) maintains dynamically evolving page relevance confidence under the current query, and \(\mathcal{M}_{\text{epi}}^{(t)}\) records intermediate findings and query refinement trajectories within the current reasoning process. These three memory states correspond to persistent document knowledge, dynamic retrieval state, and query-local reasoning experience, respectively.

\paragraph{Document Schema Memory}
\(\mathcal{M}_{\text{schema}}\) is constructed offline. The system first generates a summary \(s_i\) for each page \(p_i\), and then aggregates all page summaries into
\[
\mathcal{M}_{\text{schema}}
=
(\tau,\Sigma,\sigma),
\]
where \(\tau\) denotes the document type and topic description, \(\Sigma=\{(a_j,b_j,l_j)\}_{j=1}^{K}\) is a structural index consisting of page ranges \([a_j,b_j]\) and topic labels \(l_j\), and \(\sigma\) is a document-level global summary. This memory remains fixed during inference and serves as a query-independent navigation prior.

\paragraph{Page Belief Memory}
\(\mathcal{M}_{\text{belief}}^{(t)}\) maintains the relevance confidence of each page under the current query. For each page \(p_i\), \modelname{} uses a Beta distribution to model uncertainty over its latent relevance:
\[
\mathcal{M}_{\text{belief}}^{(t)}
=
\left\{
\operatorname{Beta}
\left(
\alpha_i^{(t)},
\beta_i^{(t)}
\right)
\right\}_{i=1}^{N}.
\]
Here, \(\alpha_i\) and \(\beta_i\) accumulate evidence supporting the page being relevant and irrelevant, respectively. To initialize this memory, we compute visual relevance between the query and each page using the late-interaction mechanism of ColQwen2.5. Let \(\mathbf{q}_k\) be the embedding of the \(k\)-th query token, and let \(\mathbf{f}_{i,j}\) be the \(j\)-th visual patch embedding of page \(p_i\). The raw retrieval score is
\[
\tilde{c}_i
=
\sum_{k=1}^{|q|}
\max_j
\cos(\mathbf{q}_k,\mathbf{f}_{i,j}).
\]
After normalization, we obtain \(c_i\in[0,1]\) and initialize the page prior as
\[
\alpha_i^{(0)} = S c_i + 1,
\qquad
\beta_i^{(0)} = S(1-c_i)+1,
\]
where \(S\) controls the strength of the visual prior. This initialization converts the first-round visual retrieval signal into a page-level prior confidence that can be continuously revised by subsequent reasoning feedback.

\paragraph{Question Episodic Memory}
\(\mathcal{M}_{\text{epi}}^{(t)}\) stores the episodic reasoning trajectory gradually formed under the current query, including discovered evidence, unmet information needs, and query rewriting history. It is updated only within a single question-answering session. Initially empty, it is written after each reasoning round as
\[
\mathcal{M}_{\text{epi}}^{(t)}
=
\{(n_k,\hat{q}_k)\}_{k=1}^{t},
\]
where \(n_k\) denotes the intermediate evidence summary and gap analysis generated by the reasoner at round \(k\), and \(\hat{q}_k\) denotes the corresponding refined query. By injecting this trajectory into subsequent retrieval rounds, \modelname{} can continue narrowing the search space along the existing evidence chain, rather than repeatedly relying on the original query for stateless retrieval.

\subsection{Bayesian Page Belief Updating}

\modelname{} treats Page Belief Memory as a retrieval memory state that evolves with reasoning feedback. After reasoning round \(t\), the reasoner outputs a set of useful pages \(U_t\) and a set of irrelevant pages \(V_t\); pages not explicitly labeled remain neutral. Based on Beta-Bernoulli conjugacy, page beliefs are updated as
\[
\alpha_i
\leftarrow
\alpha_i+\mathbb{I}[i\in U_t],
\qquad
\beta_i
\leftarrow
\beta_i+\mathbb{I}[i\in V_t].
\]
The posterior mean of page \(p_i\),
\[
\mu_i
=
\frac{\alpha_i}{\alpha_i+\beta_i},
\]
represents its accumulated relevance confidence at the current round. This state integrates the initial visual prior with cross-round reasoning feedback, and serves as the basis for subsequent retrieval control.

To exploit the locality of evidence in long documents, \modelname{} further propagates positive page feedback to neighboring pages. For \(i\in U_t\), if \(0<|j-i|\le r\), we update
\[
\alpha_j
\leftarrow
\alpha_j+\gamma^{|j-i|},
\]
where \(r\) is the propagation radius and \(\gamma\) is the distance decay factor. Updates from overlapping neighborhoods are accumulated, while negative feedback is not propagated, avoiding the suppression of potentially relevant nearby pages. Inspired by the spatial contiguity principle in cognitive load theory, this mechanism transforms local continuity in document structure into a soft prior for page exploration.

\subsection{Memory-Guided Dynamic Retrieval and Reasoning}

The updated Page Belief is used to control page selection in the next retrieval round. At round \(t\), \modelname{} samples a relevance estimate from each page posterior:
\[
\theta_i^{(t)}
\sim
\operatorname{Beta}
\left(
\alpha_i^{(t)},
\beta_i^{(t)}
\right).
\]
Meanwhile, the system recomputes the visual similarity \(c_i^{(t)}\) between each page and the current query \(\hat{q}_t\), and combines the immediate retrieval signal with the cross-round page belief into a candidate score:
\[
\operatorname{score}_i^{(t)}
=
(1-\lambda_t)c_i^{(t)}
+
\lambda_t\theta_i^{(t)}.
\]
Here, \(\lambda_t\) controls the relative weight between current-query similarity and historical page belief. As iteration proceeds, retrieval gradually shifts from relying only on visual similarity toward incorporating accumulated page confidence. Thompson sampling provides an exploration--exploitation trade-off: high-confidence pages can be repeatedly exploited, while uncertain but potentially relevant pages still retain the opportunity to be explored.

Based on the candidate scores, the system first selects a candidate page set \(\mathcal{C}_t\). A language model then reranks these candidates using their page summaries together with \(\mathcal{M}_{\text{schema}}\) and \(\mathcal{M}_{\text{epi}}^{(t)}\), producing a refined page set \(\mathcal{P}_t\). The selected pages are merged into the cross-round accumulated set:
\[
\mathcal{A}_t
=
\mathcal{A}_{t-1}\cup\mathcal{P}_t.
\]
\modelname{} then selects evidence pages from \(\mathcal{A}_t\) according to the posterior mean \(\mu_i\) and feeds them to the reasoner. Since historical pages remain in the accumulated set, previously high-confidence pages can be revisited when new evidence emerges, supporting cross-round evidence integration.

The reasoner reads the current evidence pages and memory context, and produces one of three outcomes. If the evidence is sufficient, it returns the final answer. If the document lacks adequate support, it returns \texttt{not\_answerable}. Otherwise, it generates a refined query \(\hat{q}_{t+1}\) and an intermediate note \(n_t\), triggering the next retrieval round. In this case, Question Episodic Memory is updated as
\[
\mathcal{M}_{\text{epi}}^{(t+1)}
=
\mathcal{M}_{\text{epi}}^{(t)}
\cup
\{(n_t,\hat{q}_{t+1})\}.
\]
The reasoner also outputs \(U_t\) and \(V_t\), which drive the next Page Belief update. In this way, \modelname{} continuously writes reasoning feedback into memory and transforms it into retrieval control signals, forming a retrieval--reasoning--memory update loop. The loop terminates when the system produces an answer, returns \texttt{not\_answerable}, or reaches the maximum number of rounds.

Finally, \modelname{} adopts adaptive-granularity evidence access during reasoning. For pages containing locally dense information, the system supplements the full-page context with finer-grained visual regions, allowing the reasoner to use both page-level semantics and high-resolution local evidence. This design mitigates the loss of details caused by full-page representation, especially for table-intensive documents.
\section{Experiments}
\label{sec:experiments}

\begin{table*}[t]
\centering
\small                          
\setlength{\tabcolsep}{12pt}         
\renewcommand{\arraystretch}{1.0}  

\begin{tabular}{l ccccccc}
\toprule
\multirow{2}{*}{\textbf{Method}} & \multicolumn{6}{c}{\textbf{MMLongBench}} & \multirow{2}{*}{\textbf{ALL}} \\
\cmidrule(lr){2-7}
& \textbf{TXT} & \textbf{LAY} & \textbf{CHA} & \textbf{TAB} & \textbf{FIG} & \textbf{UNA} & \\
\midrule
\rowcolor{gray!10}
\multicolumn{8}{c}{\textit{Proprietary Models}} \\
GPT-4o~\cite{hurst2024gpt} & 46.3 & 46.0 & 45.3 & 50.0 & 44.1 & 20.2 & 42.8 \\
Claude-4-Sonnet~\cite{anthropic2025system} & 50.4 & 49.4 & 50.5 & 57.3 & 43.9 & 59.0 & 53.4 \\
Gemini-2.5-Flash~\cite{comanici2025gemini} & 44.0 & 53.2 & 46.0 & 43.9 & 48.2 & 56.7 & 49.6 \\
Gemini-2.5-Pro~\cite{comanici2025gemini} & 52.1 & 62.1 & 55.5 & 55.3 & 54.0 & 59.9 & 58.1 \\
\rowcolor{gray!10}
\multicolumn{8}{c}{\textit{Open-Sourced MLLMs}} \\
Qwen3-VL-8B~\cite{bai2025qwen3} & 46.2 & 29.4 & 34.8 & 31.6 & 41.8 & 53.2 & 43.5 \\
Qwen3.5-9B~\cite{team2026qwen3} & 61.6 & 41.2 & 62.9 & 49.1 & 51.0 & 67.0 & 58.8 \\
InternVL3-8B~\cite{zhu2025internvl3} & 37.4 & 22.7 & 26.4 & 21.1 & 27.3 & 11.2 & 26.2 \\
\rowcolor{gray!10}
\multicolumn{8}{c}{\textit{Agent-based Systems}} \\
SimpleDoc~\cite{jain2025simpledoc} & 60.0 & 51.3 & 54.9 & 51.2 & 51.2 & 67.4 & 59.5 \\
MLDocRAG~\cite{zhang2026mldocrag} & 47.2 & 37.8 & 42.7 & 41.3 & 31.9 & 71.5 & 47.9 \\
AgenticOCR~\cite{wang2026agenticocr} & \underline{67.4} & \textbf{68.8} & 63.1 & 64.3 & \underline{63.6} & 66.8 & 66.4 \\
DocLens~\cite{zhu2025doclens} & 63.7 & \underline{64.6} & \underline{64.3} & \underline{69.7} & 60.2 & \underline{72.2} & \underline{67.6} \\
\midrule
\rowcolor{gray!10}
\textbf{Ours} & \textbf{70.5} & 61.3 & \textbf{67.8} & \textbf{73.3} & \textbf{63.8} & \textbf{78.8} & \textbf{71.3} \\
\bottomrule
\end{tabular}
\caption{Performance breakdown on the MMLongBench benchmark. Most baseline results are taken from the official leaderboards or the respective papers.}
\label{sec:experiment}
\label{tab:mmlong_results}
\end{table*}

\begin{table}[t]
    \centering
    \resizebox{\columnwidth}{!}{
    \begin{tabular}{l c c c c}
        \toprule
        \textbf{Method} & \textbf{Model} & 
        \textbf{Retriever} &
        \textbf{Rounds} & \textbf{ALL} \\
        \midrule
        
        MoLoRAG & Qwen3-VL-8B & ColQwen2.5 & - & 55.0 \\
        SimpleDoc & Qwen3-VL-8B & ColQwen2.5 & Iter 3 & 60.1 \\
        \rowcolor{gray!10}
        \textbf{Ours} & Qwen3-VL-8B & ColQwen2.5 & Iter 3 & \textbf{61.7} \\
        
        \midrule
        
        MoLoRAG & Qwen3.5-9B & ColQwen2.5 & - & 67.1 \\
        SimpleDoc & Qwen3.5-9B & ColQwen2.5 & Iter 3 & 69.3 \\
        \rowcolor{gray!10}
        \textbf{Ours} & Qwen3.5-9B & ColQwen2.5 & Iter 3 & \textbf{71.3} \\
        
        \bottomrule
    \end{tabular}
    }
    \caption{Comparison under the same configuration.}
    \label{tab:vlm_comparison}
\end{table}

\subsection{Experimental Setup}

\paragraph{Benchmarks}
We evaluate \modelname{} on three long-document DocVQA benchmarks, covering diverse document types, evidence modalities, and reasoning requirements. MMLongBench-Doc~\cite{ma2024mmlongbench} contains 1,082 questions over 135 long documents, with documents spanning up to 112 pages and averaging 47.5 pages. Its questions cover multiple evidence types, including text, images, tables, charts, layout understanding, and unanswerable cases (UNA). LongDocURL~\cite{deng2025longdocurl} consists of 2,325 question-answer pairs over 396 PDF documents, targeting long-document understanding, numerical reasoning, and cross-element grounding. PaperTab~\cite{hui2024uda} provides 393 questions over 307 scientific papers, with a focus on retrieving, interpreting, and reasoning over tabular information in academic documents. Overall, these benchmarks span a broad range of evidence types and task complexities.

\paragraph{Evaluation protocol}
Following recent baselines, we use Accuracy as the primary evaluation metric and adopt GPT-4.1 as an automatic binary judge to determine whether each response is correct or incorrect. Specifically, each prediction receives a score in ${0,1}$, and Accuracy is computed as the proportion of correctly answered questions. To analyze iterative retrieval behavior, we further report Evidence Recall, defined as the proportion of ground-truth evidence pages covered by the accumulated retrieved pages, and All-Hit Rate, defined as the fraction of questions for which all annotated evidence pages are successfully retrieved.

\begin{table}[t]
    \centering
    
    \resizebox{\columnwidth}{!}{ 
        \begin{tabular}{l cccc}
            \toprule
            \textbf{Method} & \textbf{MMLong} & \textbf{LongDoc} & \textbf{PaperTab} & \textbf{Avg.} \\
            \midrule
            GPT-4o & 44.9 & 64.5 & 40.3 & 49.9 \\
            Qwen3-VL-8B & 43.5 & 37.0 & 58.3 & 46.3 \\
            Qwen3.5-9B & 58.8 & 48.0 & \underline{68.7} & 58.5 \\
            InternVL3-8B & 26.2 & 30.5 & 43.8 & 33.5 \\
            LLaMA3.1-8B & 22.6 & 29.8 & 13.5 & 22.0 \\
            SimpleDoc & \underline{60.6} & \underline{72.3} & 65.4 & \underline{66.1} \\
            MoLoRAG & 41.0 & 51.9 & 32.3 & 41.7 \\
            MLDocRAG & 47.9 & 50.8 & - & 49.4 \\
            \midrule
            \rowcolor{gray!10}
            \textbf{Ours} & \textbf{71.3} & \textbf{81.1} & \textbf{80.4} & \textbf{77.6} \\
            \bottomrule
        \end{tabular}
    } 
    \caption{Overall results on MMLongBench-Doc, LongDocURL, and PaperTab.}
    \label{tab:overall_single}
\end{table}

\paragraph{Implementation details}
All experiments are conducted on NVIDIA A100 GPUs. \modelname{} uses Qwen3.5-VL-9B as the backbone model. For visual retrieval, we employ ColQwen2.5-v0.2 ~\cite{faysse2025colpali} to offline encode PDF pages. The backbone model is served with vLLM; page-level textual summaries are generated offline and stored together with visual embeddings. Structure-aware table and figure crops are extracted using MinerU ~\cite{wang2024mineru}. For Page Belief Memory, we set the Beta prior initialization strength to $S{=}5.0$; the cross-round fusion schedule to $\lambda{=}[0.0, 0.3, 0.6, 0.6]$; the spatial propagation radius to $r{=}2$; and the decay factor to $\gamma{=}0.5$. Additional hyperparameters and implementation details are provided in the appendix.

\begin{table*}[t]
    \centering
    \setlength{\tabcolsep}{8pt}
    \begin{tabular}{l ccccccc}
            \toprule
            \textbf{Method} & \textbf{TXT} & \textbf{LAY} & \textbf{CHA} & \textbf{TAB} & \textbf{FIG} & \textbf{UNA} & \textbf{ALL} \\
            \midrule
            (w/o) Document Schema Memory & 70.37 & 60.86 & 63.84 & 69.91 & 62.17 & 78.63 & 70.16 \\
            (w/o) Page Belief Memory & 68.09 & 53.78 & 61.93 & 70.18 & 57.89 & 76.96 & 68.80 \\
            (w/o) Question Episodic Memory & 69.41 & 60.50 & 60.45 & 70.37 & 61.18 & 74.78 & 69.02 \\
            (w/o) All Memory Modules & 67.41 & 61.34 & 62.15 & 70.18 & 63.12 & 78.56 & 68.47 \\
            (w/o) Bayesian updating & 67.75 & 52.94 & 63.07 & 69.27 & 57.89 & 76.96 & 68.80 \\
            (w/o) Thompson sampling & 70.25 & 60.17 & 66.09 & 70.09 & 61.74 & 78.79 & 69.89 \\
            (w/o) Adaptive-granularity & 68.23 & 58.82 & 64.20 & 68.66 & 63.16 & 78.79 & 69.91 \\
            \midrule
            \rowcolor{gray!10}
            \textbf{Ours (Full)} & \textbf{70.53} & \textbf{61.34} & \textbf{67.80} & \textbf{73.27} & \textbf{63.77} & \textbf{78.79} & \textbf{71.28} \\
            \bottomrule
        \end{tabular}
    \caption{Ablation study on different modules.}
    \label{tab:ablation_study}
\end{table*}

\subsection{Comparison with State-of-the-arts}
\paragraph{Overall Performance}
Baseline methods follow the standard protocols of the corresponding leaderboards and prior work. Table~\ref{tab:overall_single} summarizes the overall performance of \modelname{} on three long-document DocVQA benchmarks. The compared methods cover proprietary multimodal models, open-source multimodal large language models, and agentic systems for long-document understanding. Relative to the strongest agentic baselines on each benchmark, \modelname{} improves performance by 3.7 and 8.8 percentage points on MMLongBench-Doc and LongDocURL, respectively. On PaperTab, \modelname{} yields a further 15.0-point gain, highlighting the benefit of Bayesian page belief updating and adaptive-granularity parsing for evidence-dense documents.

\paragraph{Fine-grained Analysis on MMLongBench-Doc}
Table~\ref{sec:experiment} further reports category-wise results on MMLongBench-Doc. \modelname{} achieves an overall accuracy of 71.3\% and attains the best performance. Its gains are especially pronounced on TAB and UNA, reaching 73.3\% and 78.8\%, respectively. These results indicate that page belief updating together with fine-grained table evidence augmentation improves evidence localization in table-intensive settings, while accumulated cross-round retrieval states help the system more reliably identify truly unanswerable questions. 

To further demonstrate the effectiveness of \modelname{}, we conduct controlled experiments against recent open-source baselines under the same backbone model, retriever, maximum page budget, and maximum retrieval rounds. 
All methods are evaluated using GPT-4.1 with identical evaluation prompts, and the results are summarized in Table~\ref{tab:vlm_comparison}. \modelname{} consistently achieves the best performance under this controlled setting. Additional analyses, including key hyperparameter sensitivity, human verification of automatic evaluation results, and stronger theoretical justifications, are provided in the Appendix.

\begin{figure}[t]
    \centering
    \includegraphics[width=\columnwidth]{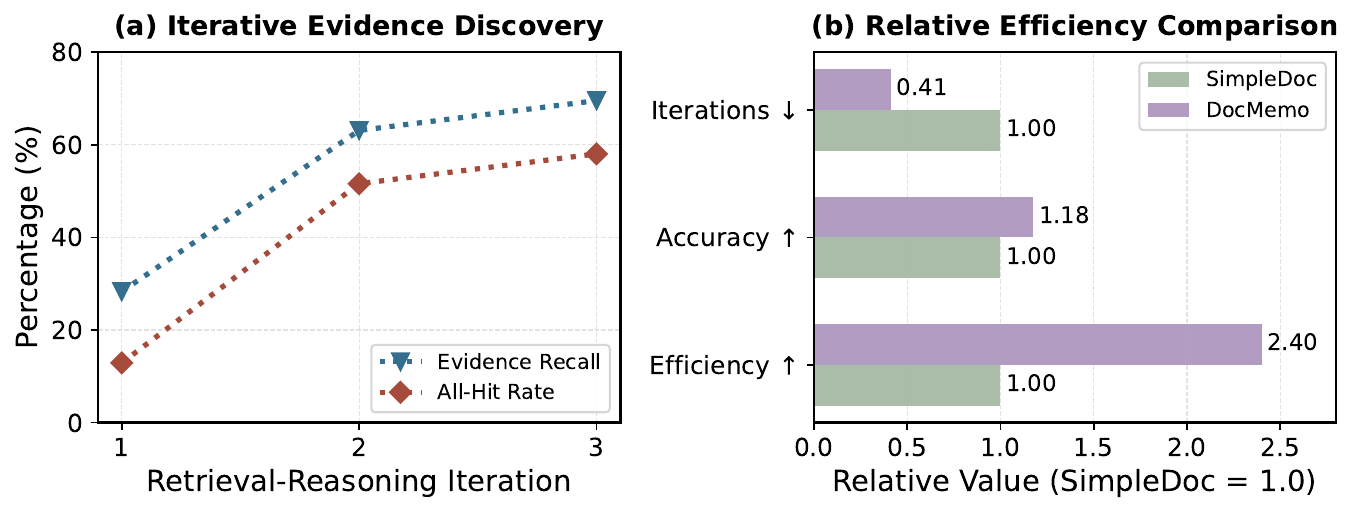}
    \caption{
    Iterative evidence discovery and relative efficiency comparison.
    (a) Evidence recall and all-hit rate progressively improve with iterative retrieval-reasoning.
    (b) Relative efficiency comparison.
    }
    \label{fig:efficiency}
\end{figure}

\subsection{Ablation Studies}
Table~\ref{tab:ablation_study} reports the ablation results of \modelname{} across different evidence types on MMLongBench-Doc. Removing either the tri-level memory or Bayesian page belief updating leads to a clear drop in overall accuracy, from 71.3\% to 68.5\% and 68.8\%, respectively. The degradation is particularly evident on major categories such as CHA and TXT, suggesting that when evidence is dispersed or requires cross-page clue integration, relying only on the retrieval signal from the current round is insufficient for robust reasoning. Since Page Belief Memory and Question Episodic Memory are dynamically updated online and directly participate in cross-round information propagation, they play a more critical role in improving the overall performance.

Disabling adaptive-granularity evidence access leads to the largest degradation on TAB, while its impact on other categories is comparatively smaller. Overall, the ablations confirm that \modelname{} benefits from both structured retrieval memory and dynamic belief revision, with fine-grained evidence augmentation providing additional gains in visually dense regions.

\begin{figure*}[t]
    \centering
    \includegraphics[width=0.95\textwidth]{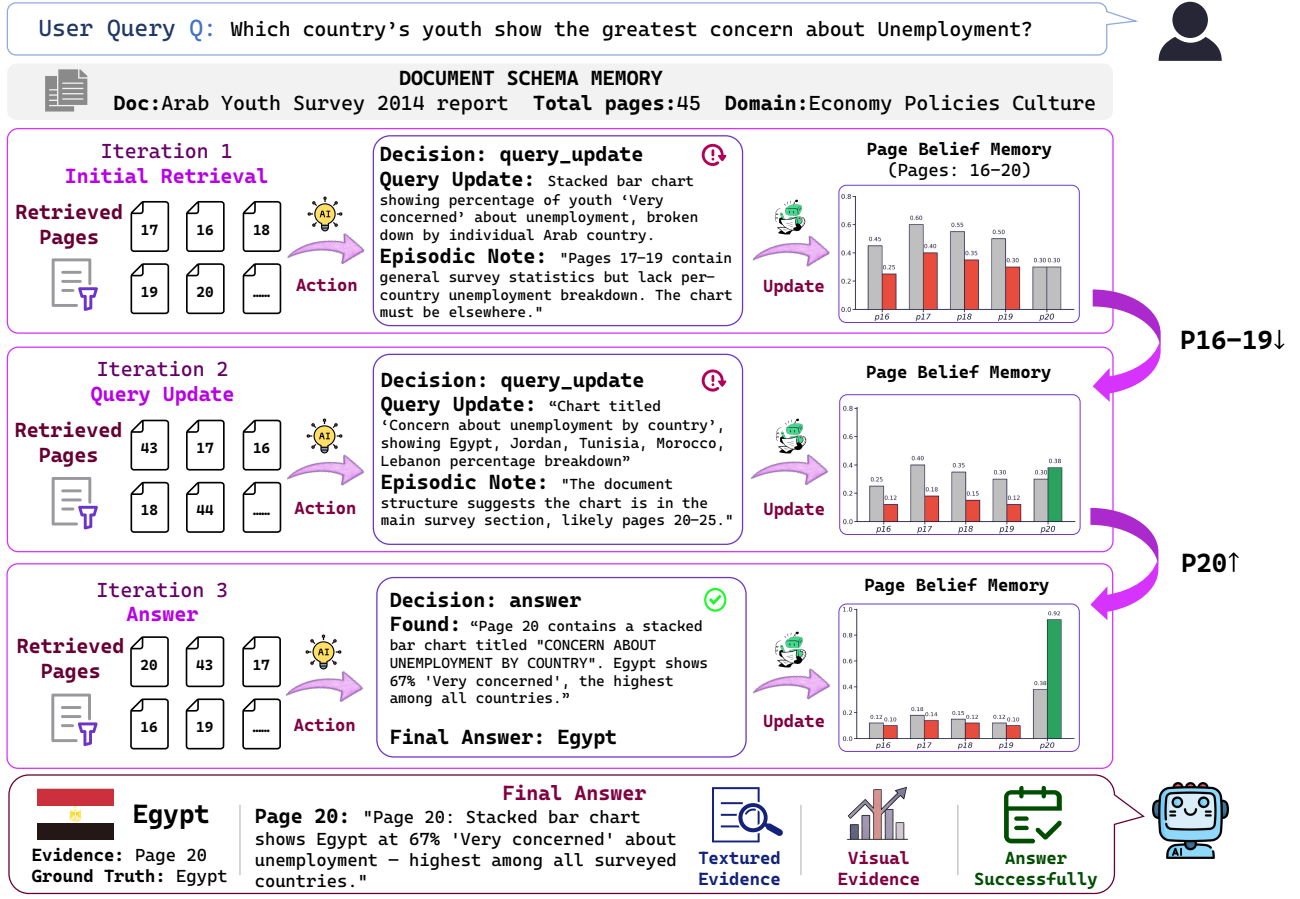}
    
    \caption{A qualitative example of \modelname{}'s memory-guided iterative retrieval. }
    \label{fig:ablation_double}
\end{figure*}

\subsection{Iterative Retrieval and Efficiency Analysis}
To analyze \modelname{}'s evidence discovery ability during multi-round retrieval, Figure~\ref{fig:efficiency}(a) illustrates the evolution of cumulative evidence recall over retrieval--reasoning iterations. Evidence Recall improves from 28.32\% in the first round to 69.56\% in the third round, while All-Hit Rate increases from 12.90\% to 58.05\%, demonstrating that cross-round memory states and page-level confidence updates can continuously recover missing evidence. The largest improvement occurs in the second iteration, indicating that reasoning feedback effectively guides subsequent evidence exploration, while the third iteration provides diminishing gains. Furthermore, Figure~\ref{fig:efficiency}(b) compares the retrieval efficiency between \modelname{} and SimpleDoc. \modelname{} reduces the number of iterations to 0.41$\times$ while achieving higher accuracy (1.18$\times$), resulting in a 2.40$\times$ overall efficiency improvement. This indicates that memory-guided dynamic retrieval can reduce unnecessary searches and improve evidence acquisition efficiency. In addition, the Appendix provides further supplementary experiments, including detailed analyses of iteration efficiency and offline/online inference costs. 

\subsection{Qualitative Analysis}
\label{sec:qualitative}
Figure~\ref{fig:ablation_double} illustrates how structured memory, dynamic page belief updating, and adaptive evidence access work together in long-document reasoning. By writing reasoning feedback from retrieved pages into Page Belief Memory, \modelname{} gradually shifts retrieval from initial static similarity matching toward dynamic exploration guided by cross-round evidence states. As Question Episodic Memory accumulates the missing information under the current query, subsequent queries become more targeted toward uncovered evidence types, while Bayesian belief updating and spatial propagation further increase the priority of potentially relevant pages. Overall, the iterative process in \modelname{} is not merely about increasing the number of retrieval rounds; rather, it continuously reshapes the search space through structured memory, enabling the system to recover from insufficient early evidence and locate the key chart page.
\section{Conclusion}
\label{sec:conclusion}

We present \modelname{}, a memory-guided iterative reasoning framework for long-document visual question answering. \modelname{} explicitly disentangles document structural priors, page-level relevance confidence, and query-specific reasoning trajectories through a tri-level document memory. It further uses Bayesian page belief updating and Thompson sampling to continuously transform cross-round reasoning feedback into retrieval signals. Combined with adaptive-granularity evidence access, \modelname{} enhances fine-grained reading over information-dense page regions. Experiments on three long-document DocVQA benchmarks show that \modelname{} consistently improves over static retrieval and iterative baselines, validating the effectiveness of structured memory and dynamic evidence exploration.

\bibliography{main}

@article{ma2024mmlongbench,
  title={Mmlongbench-doc: Benchmarking long-context document understanding with visualizations},
  author={Ma, Yubo and Zang, Yuhang and Chen, Liangyu and Chen, Meiqi and Jiao, Yizhu and Li, Xinze and Lu, Xinyuan and Liu, Ziyu and Ma, Yan and Dong, Xiaoyi and others},
  journal={Advances in Neural Information Processing Systems},
  volume={37},
  pages={95963--96010},
  year={2024}
}

@inproceedings{deng2025longdocurl,
  title={Longdocurl: a comprehensive multimodal long document benchmark integrating understanding, reasoning, and locating},
  author={Deng, Chao and Yuan, Jiale and Bu, Pi and Wang, Peijie and Li, Zhong-Zhi and Xu, Jian and Li, Xiao-Hui and Gao, Yuan and Song, Jun and Zheng, Bo and others},
  booktitle={Proceedings of the 63rd Annual Meeting of the Association for Computational Linguistics (Volume 1: Long Papers)},
  pages={1135--1159},
  year={2025}
}

@article{hui2024uda,
  title={Uda: A benchmark suite for retrieval augmented generation in real-world document analysis},
  author={Hui, Yulong and Lu, Yao and Zhang, Huanchen},
  journal={Advances in Neural Information Processing Systems},
  volume={37},
  pages={67200--67217},
  year={2024}
}

@inproceedings{jain2025simpledoc,
  title={SimpleDoc: Multi-Modal Document Understanding with Dual-Cue Page Retrieval and Iterative Refinement},
  author={Jain, Chelsi and Wu, Yiran and Zeng, Yifan and Liu, Jiale and Dai, Shengyu and Shao, Zhenwen and Wu, Qingyun and Wang, Huazheng},
  booktitle={Proceedings of the 2025 Conference on Empirical Methods in Natural Language Processing},
  pages={28398--28415},
  year={2025}
}

@article{lewis2020retrieval,
  title={Retrieval-augmented generation for knowledge-intensive nlp tasks},
  author={Lewis, Patrick and Perez, Ethan and Piktus, Aleksandra and Petroni, Fabio and Karpukhin, Vladimir and Goyal, Naman and K{\"u}ttler, Heinrich and Lewis, Mike and Yih, Wen-tau and Rockt{\"a}schel, Tim and others},
  journal={Advances in neural information processing systems},
  volume={33},
  pages={9459--9474},
  year={2020}
}

@inproceedings{karpukhin2020dense,
  title={Dense passage retrieval for open-domain question answering},
  author={Karpukhin, Vladimir and Oguz, Barlas and Min, Sewon and Lewis, Patrick and Wu, Ledell and Edunov, Sergey and Chen, Danqi and Yih, Wen-tau},
  booktitle={Proceedings of the 2020 conference on empirical methods in natural language processing (EMNLP)},
  pages={6769--6781},
  year={2020}
}

@article{zhu2025doclens,
  title={Doclens: A tool-augmented multi-agent framework for long visual document understanding},
  author={Zhu, Dawei and Meng, Rui and Chen, Jiefeng and Li, Sujian and Pfister, Tomas and Yoon, Jinsung},
  journal={arXiv preprint arXiv:2511.11552},
  year={2025}
}

@inproceedings{tanaka2025vdocrag,
  title={Vdocrag: Retrieval-augmented generation over visually-rich documents},
  author={Tanaka, Ryota and Iki, Taichi and Hasegawa, Taku and Nishida, Kyosuke and Saito, Kuniko and Suzuki, Jun},
  booktitle={Proceedings of the Computer Vision and Pattern Recognition Conference},
  pages={24827--24837},
  year={2025}
}

@inproceedings{yu2025visrag,
  title={Visrag: Vision-based retrieval-augmented generation on multi-modality documents},
  author={Yu, Shi and Tang, Chaoyue and Xu, Bokai and Cui, Junbo and Ran, Junhao and Yan, Yukun and Liu, Zhenghao and Wang, Shuo and Han, Xu and Liu, Zhiyuan and others},
  booktitle={International Conference on Learning Representations},
  volume={2025},
  pages={21074--21098},
  year={2025}
}

@article{cho2024m3docrag,
  title={M3docrag: Multi-modal retrieval is what you need for multi-page multi-document understanding},
  author={Cho, Jaemin and Mahata, Debanjan and Irsoy, Ozan and He, Yujie and Bansal, Mohit},
  journal={arXiv preprint arXiv:2411.04952},
  year={2024}
}

@inproceedings{asai2024self,
  title={Self-rag: Learning to retrieve, generate, and critique through self-reflection},
  author={Asai, Akari and Wu, Zeqiu and Wang, Yizhong and Sil, Avi and Hajishirzi, Hannaneh},
  booktitle={International conference on learning representations},
  volume={2024},
  pages={9112--9141},
  year={2024}
}

@inproceedings{jiang2023active,
  title={Active retrieval augmented generation},
  author={Jiang, Zhengbao and Xu, Frank F and Gao, Luyu and Sun, Zhiqing and Liu, Qian and Dwivedi-Yu, Jane and Yang, Yiming and Callan, Jamie and Neubig, Graham},
  booktitle={Proceedings of the 2023 conference on empirical methods in natural language processing},
  pages={7969--7992},
  year={2023}
}

@inproceedings{trivedi2023interleaving,
  title={Interleaving retrieval with chain-of-thought reasoning for knowledge-intensive multi-step questions},
  author={Trivedi, Harsh and Balasubramanian, Niranjan and Khot, Tushar and Sabharwal, Ashish},
  booktitle={Proceedings of the 61st annual meeting of the association for computational linguistics (volume 1: long papers)},
  pages={10014--10037},
  year={2023}
}

@article{mcclelland1995there,
  title={Why there are complementary learning systems in the hippocampus and neocortex: insights from the successes and failures of connectionist models of learning and memory.},
  author={McClelland, James L and McNaughton, Bruce L and O'Reilly, Randall C},
  journal={Psychological review},
  volume={102},
  number={3},
  pages={419},
  year={1995},
  publisher={American Psychological Association}
}

@article{dong2025towards,
  title={Towards large language models with human-like episodic memory},
  author={Dong, Cody V and Lu, Qihong and Norman, Kenneth A and Michelmann, Sebastian},
  journal={Trends in Cognitive Sciences},
  year={2025},
  publisher={Elsevier}
}

@article{xu2026mem,
  title={A-mem: Agentic memory for llm agents},
  author={Xu, Wujiang and Liang, Zujie and Mei, Kai and Gao, Hang and Tan, Juntao and Zhang, Yongfeng},
  journal={Advances in Neural Information Processing Systems},
  volume={38},
  pages={17577--17604},
  year={2026}
}

@article{yan2025memory,
  title={Memory-r1: Enhancing large language model agents to manage and utilize memories via reinforcement learning},
  author={Yan, Sikuan and Yang, Xiufeng and Huang, Zuchao and Nie, Ercong and Ding, Zifeng and Li, Zonggen and Ma, Xiaowen and Bi, Jinhe and Kersting, Kristian and Pan, Jeff Z and others},
  journal={arXiv preprint arXiv:2508.19828},
  year={2025}
}

@article{thompson1933likelihood,
  title={On the likelihood that one unknown probability exceeds another in view of the evidence of two samples},
  author={Thompson, William R},
  journal={Biometrika},
  volume={25},
  number={3/4},
  pages={285--294},
  year={1933},
  publisher={JSTOR}
}

@article{sweller1988cognitive,
  title={Cognitive load during problem solving: Effects on learning},
  author={Sweller, John},
  journal={Cognitive science},
  volume={12},
  number={2},
  pages={257--285},
  year={1988},
  publisher={Elsevier}
}

@article{wang2026agenticocr,
  title={Agenticocr: Parsing only what you need for efficient retrieval-augmented generation},
  author={Wang, Zhengren and Ma, Dongsheng and Zhong, Huaping and Li, Jiayu and Zhang, Wentao and Wang, Bin and He, Conghui},
  journal={arXiv preprint arXiv:2602.24134},
  year={2026}
}

@inproceedings{li2026regionrag,
  title={RegionRAG: Region-level Retrieval-Augmented Generation for Visual Document Understanding},
  author={Li, Yinglu and Lu, Zhiying and Liu, Zhihang and Sun, Yiwei and Liu, Chuanbin and Xie, Hongtao},
  booktitle={Proceedings of the AAAI Conference on Artificial Intelligence},
  volume={40},
  number={8},
  pages={6662--6670},
  year={2026}
}

@article{gao2023retrieval,
  title={Retrieval-augmented generation for large language models: A survey},
  author={Gao, Yunfan and Xiong, Yun and Gao, Xinyu and Jia, Kangxiang and Pan, Jinliu and Bi, Yuxi and Dai, Yixin and Sun, Jiawei and Wang, Haofen and Wang, Haofen and others},
  journal={arXiv preprint arXiv:2312.10997},
  volume={2},
  number={1},
  pages={32},
  year={2023}
}

@inproceedings{wu2025molorag,
  title={Molorag: Bootstrapping document understanding via multi-modal logic-aware retrieval},
  author={Wu, Xixi and Tan, Yanchao and Hou, Nan and Zhang, Ruiyang and Cheng, Hong},
  booktitle={Proceedings of the 2025 Conference on Empirical Methods in Natural Language Processing},
  pages={14035--14056},
  year={2025}
}

@inproceedings{liu2024ra,
  title={Ra-isf: Learning to answer and understand from retrieval augmentation via iterative self-feedback},
  author={Liu, Yanming and Peng, Xinyue and Zhang, Xuhong and Liu, Weihao and Yin, Jianwei and Cao, Jiannan and Du, Tianyu},
  booktitle={Findings of the Association for Computational Linguistics: ACL 2024},
  pages={4730--4749},
  year={2024}
}

@article{packer2023memgpt,
  title={MemGPT: towards LLMs as operating systems.},
  author={Packer, Charles and Fang, Vivian and Patil, Shishir\_G and Lin, Kevin and Wooders, Sarah and Gonzalez, Joseph\_E},
  year={2023},
  publisher={ArXiv}
}

@inproceedings{he2024ma,
  title={Ma-lmm: Memory-augmented large multimodal model for long-term video understanding},
  author={He, Bo and Li, Hengduo and Jang, Young Kyun and Jia, Menglin and Cao, Xuefei and Shah, Ashish and Shrivastava, Abhinav and Lim, Ser-Nam},
  booktitle={Proceedings of the IEEE/CVF conference on computer vision and pattern recognition},
  pages={13504--13514},
  year={2024}
}

@article{long2025seeing,
  title={Seeing, listening, remembering, and reasoning: A multimodal agent with long-term memory},
  author={Long, Lin and He, Yichen and Ye, Wentao and Pan, Yiyuan and Lin, Yuan and Li, Hang and Zhao, Junbo and Li, Wei},
  journal={arXiv preprint arXiv:2508.09736},
  year={2025}
}

@article{huang2025licomemory,
  title={Licomemory: Lightweight and cognitive agentic memory for efficient long-term reasoning},
  author={Huang, Zhengjun and Tian, Zhoujin and Guo, Qintian and Zhang, Fangyuan and Zhou, Yingli and Jiang, Di and Xie, Zeying and Zhou, Xiaofang},
  journal={arXiv preprint arXiv:2511.01448},
  year={2025}
}

@article{lian2026verbatim,
  title={From Verbatim to Gist: Distilling Pyramidal Multimodal Memory via Semantic Information Bottleneck for Long-Horizon Video Agents},
  author={Lian, Niu and Wang, Yuting and Yao, Hanshu and Wang, Jinpeng and Chen, Bin and Wang, Yaowei and Zhang, Min and Xia, Shu-Tao},
  journal={arXiv preprint arXiv:2603.01455},
  year={2026}
}

@article{xie2024large,
  title={Large multimodal agents: A survey},
  author={Xie, Junlin and Chen, Zhihong and Zhang, Ruifei and Wan, Xiang and Li, Guanbin},
  journal={arXiv preprint arXiv:2402.15116},
  year={2024}
}

@article{zhang2024document,
  title={Document parsing unveiled: Techniques, challenges, and prospects for structured information extraction},
  author={Zhang, Qintong and Wang, Bin and Huang, Victor Shea-Jay and Zhang, Junyuan and Wang, Zhengren and Liang, Hao and He, Conghui and Zhang, Wentao},
  journal={arXiv preprint arXiv:2410.21169},
  year={2024}
}

@article{wang2024mineru,
  title={Mineru: An open-source solution for precise document content extraction},
  author={Wang, Bin and Xu, Chao and Zhao, Xiaomeng and Ouyang, Linke and Wu, Fan and Zhao, Zhiyuan and Xu, Rui and Liu, Kaiwen and Qu, Yuan and Shang, Fukai and others},
  journal={arXiv preprint arXiv:2409.18839},
  year={2024}
}

@article{hurst2024gpt,
  title={Gpt-4o system card},
  author={Hurst, Aaron and Lerer, Adam and Goucher, Adam P and Perelman, Adam and Ramesh, Aditya and Clark, Aidan and Ostrow, AJ and Welihinda, Akila and Hayes, Alan and Radford, Alec and others},
  journal={arXiv preprint arXiv:2410.21276},
  year={2024}
}

@article{comanici2025gemini,
  title={Gemini 2.5: Pushing the frontier with advanced reasoning, multimodality, long context, and next generation agentic capabilities},
  author={Comanici, Gheorghe and Bieber, Eric and Schaekermann, Mike and Pasupat, Ice and Sachdeva, Noveen and Dhillon, Inderjit and Blistein, Marcel and Ram, Ori and Zhang, Dan and Rosen, Evan and others},
  journal={arXiv preprint arXiv:2507.06261},
  year={2025}
}

@article{bai2025qwen3,
  title={Qwen3-vl technical report},
  author={Bai, Shuai and Cai, Yuxuan and Chen, Ruizhe and Chen, Keqin and Chen, Xionghui and Cheng, Zesen and Deng, Lianghao and Ding, Wei and Gao, Chang and Ge, Chunjiang and others},
  journal={arXiv preprint arXiv:2511.21631},
  year={2025}
}

@article{team2026qwen3,
  title={Qwen3. 5-omni technical report},
  author={Team, Qwen},
  journal={arXiv preprint arXiv:2604.15804},
  year={2026}
}

@article{zhu2025internvl3,
  title={Internvl3: Exploring advanced training and test-time recipes for open-source multimodal models},
  author={Zhu, Jinguo and Wang, Weiyun and Chen, Zhe and Liu, Zhaoyang and Ye, Shenglong and Gu, Lixin and Tian, Hao and Duan, Yuchen and Su, Weijie and Shao, Jie and others},
  journal={arXiv preprint arXiv:2504.10479},
  year={2025}
}

@article{zhang2026mldocrag,
  title={MLDocRAG: Multimodal Long-Context Document Retrieval Augmented Generation},
  author={Zhang, Yongyue and Wu, Yaxiong},
  journal={arXiv preprint arXiv:2602.10271},
  year={2026}
}

@article{anthropic2025system,
  title={System card: Claude opus 4 \& claude sonnet 4},
  author={Anthropic, AI},
  journal={Claude-4 Model Card},
  year={2025}
}

@inproceedings{faysse2025colpali,
  title={Colpali: Efficient document retrieval with vision language models},
  author={Faysse, Manuel and Sibille, Hugues and Wu, Tony and Omrani, Bilel and Viaud, Gautier and Hudelot, C{\'e}line and Colombo, Pierre},
  booktitle={International Conference on Learning Representations},
  volume={2025},
  pages={61424--61449},
  year={2025}
}



\clearpage
\appendix
\section{Hyperparameter Settings}

\begin{table}[h]
\centering
\small
\begin{tabular}{ll}
\noalign{\hrule height 1.2pt}

\multicolumn{2}{c}{Page Belief Memory} \\ \hline
Prior strength $S$                         & 5.0 \\
Thompson sampling schedule $\lambda_t$     & [0.0, 0.3, 0.6, 0.6] \\
Spatial propagation radius $r$             & 2 \\
Spatial decay factor $\gamma$              & 0.5 \\ \hline\hline

\multicolumn{2}{c}{Iterative Retrieval} \\ \hline
Max retrieval--reasoning cycles $T$        & 3 \\
Candidate pool size $K_c$                  & 30 \\
Max LLM-selected pages $K_r$               & 13 \\
Max pages per VLM call $K_q$               & 9 \\ \hline\hline

\multicolumn{2}{c}{Inference} \\ \hline
Page rendering DPI                         & 150 \\
Max tokens for retrieval LLM               & 8{,}192 \\
Max tokens for reasoning VLM               & 2{,}048 \\
Max text chars per page                    & $(32768 / 2 \times 4) / K_q$ \\ \hline\hline

\multicolumn{2}{c}{Adaptive-Granularity Evidence Access} \\ \hline
Max table crops per query                  & 5 \\
Max image long side                        & 1{,}500 px \\ \hline\hline

\multicolumn{2}{c}{Models} \\ \hline
Visual embedder                            & ColQwen2.5-v0.2 \\
Retrieval LLM                              & Qwen3.5-9B \\
Reasoning VLM                              & Qwen3.5-9B \\
Evaluation judge                           & GPT-4.1 \\ \hline

\noalign{\hrule height 1.2pt}
\end{tabular}
\caption{Hyperparameters used in \textsc{DocMemo}.}
\label{tab:hyperparams}
\end{table}

\begin{table}[h]
\centering
\resizebox{\columnwidth}{!}{%
\begin{tabular}{c ccccccc}
\toprule
\textbf{$S$} & \textbf{TXT} & \textbf{LAY} & \textbf{CHA} & \textbf{TAB} & \textbf{FIG} & \textbf{UNA} & \textbf{ALL} \\
\midrule
1  & 72.28 & 62.39 & 65.71 & 66.82 & 61.87 & 74.89 & 69.37 \\
5  & 70.53 & 61.34 & 67.80 & 73.27 & 63.77 & 78.79 & 71.28 \\
10 & 71.43 & 60.17 & 66.29 & 74.18 & 59.74 & 80.45 & 70.79 \\
\bottomrule
\end{tabular}%
}
\caption{Sensitivity analysis of the prior strength $S$.}
\label{tab:sensitivity_S}
\end{table}

\begin{table}[h]
\centering
\resizebox{\columnwidth}{!}{%
\begin{tabular}{c ccccccc}
\toprule
\textbf{$\lambda$} & \textbf{TXT} & \textbf{LAY} & \textbf{CHA} & \textbf{TAB} & \textbf{FIG} & \textbf{UNA} & \textbf{ALL} \\
\midrule
0.3, 0.3, 0.3, 0.3 & 70.20 & 63.16 & 63.22 & 72.30 & 62.63 & 78.73 & 70.66 \\
0.0, 0.3, 0.6, 0.6 & 70.53 & 61.34 & 67.80 & 73.27 & 63.77 & 78.79 & 71.28 \\
0.6, 0.6, 0.3, 0.0 & 70.29 & 58.47 & 62.15 & 69.44 & 60.60 & 76.34 & 69.04 \\
0.6, 0.6, 0.6, 0.6 & 70.20 & 59.66 & 62.71 & 72.09 & 60.60 & 77.26 & 69.88 \\
\bottomrule
\end{tabular}%
}
\caption{Sensitivity analysis of the weight parameters $\lambda$.}
\label{tab:sensitivity_lambda}
\end{table}

The hyperparameters used in our experimental setup are shown in Table~\ref{tab:hyperparams}. 
For Page Belief Memory, the prior strength \(S\) controls the influence of the initial visual retrieval score on the Beta prior, while \(\lambda_t\) determines the interpolation between current visual similarity and sampled page belief across retrieval rounds. 
The spatial propagation radius \(r\) and decay factor \(\gamma\) control how positive feedback from useful pages is propagated to neighboring pages. 
During iterative retrieval, \textsc{DocMemo} first retains \(K_c=30\) candidate pages, then uses the LLM reranker to select at most \(K_r=13\) pages, and finally feeds up to \(K_q=9\) pages to the reasoning VLM according to posterior page confidence. 
For adaptive-granularity evidence access, at most five table crops are injected per query, with the longest image side limited to 1{,}500 pixels to control visual input cost. 
Here, \(T=3\) denotes three complete Retriever--Reasoner cycles.

We conducted additional sensitivity analyses on $S$ and $\lambda$ on the full MMLongBench test set. As shown in Table~\ref{tab:sensitivity_S} and Table~\ref{tab:sensitivity_lambda}, our method remains relatively stable under different settings. Moreover, the default increasing schedule outperforms the reverse schedule. This indicates that progressively increasing the reliance on the Bayesian posterior as multi-round feedback accumulates is a more reasonable design. The settings of $r$ and $\gamma$ are mainly motivated by empirical observations that document content is often distributed continuously across neighboring pages: $r = 2$ is used to cover locally relevant pages while avoiding cross-section noise, and $\gamma = 0.5$ allows the propagation effect to gradually decay with distance, thereby limiting error propagation.

\section{Details of Prompts}
\noindent
\modelname{} inherits the basic structured summarization and retrieval prompt format from SimpleDoc, and augments it with memory-oriented output constraints required by tri-level document memory and page belief updating. 
While SimpleDoc primarily relies on page summaries, candidate-page retrieval, and reasoner-side query updates, \modelname{} further exposes intermediate states that can be explicitly written into memory: document structural priors are used to construct \textit{Document Schema Memory}; \texttt{useful\_pages} and \texttt{irrelevant\_pages} provide feedback for Bayesian updates in \textit{Page Belief Memory}; and \texttt{query\_update} and \texttt{notes} are accumulated in \textit{Question Episodic Memory}. 
The reasoning prompt also supports full-page images, extracted text, and high-resolution local regions, enabling adaptive-granularity evidence access. 
This section presents the key prompt fragments that are directly related to the proposed mechanisms.

\noindent\textbf{Document schema construction}
\modelname{} first constructs query-independent document structural memory from VLM-generated page-level summaries. 
The prompt asks the model to produce three complementary fields: \texttt{doc\_type}, which describes the document genre and topic; \texttt{doc\_structure}, which summarizes semantic sections with page ranges; and \texttt{doc\_summary}, which provides a global document summary. 
This structured representation serves as a document-level navigation prior for subsequent retrieval. 
For long documents, we adopt a batch-and-merge strategy: the system first identifies thematic transitions within local page segments and then merges them into a unified document structure.

\begin{lstlisting}[language={}, caption={Prompt excerpt for constructing Document Schema Memory.}]
Your analysis must produce three outputs:

1. Document Type (doc_type)
Provide a specific, free-form description of the document's genre
and subject matter.

2. Document Structure (doc_structure)
Identify the major thematic sections of the document with their
approximate page ranges.

3. Document Summary (doc_summary)
Write 4-6 sentences that describe what the document covers overall.

Output strictly in the following format:

<doc_type>
[specific description of document genre and topic]
</doc_type>

<doc_structure>
Pages X-Y: [section topic]
Pages Y-Z: [section topic]
...
</doc_structure>

<doc_summary>
[4-6 sentences describing the document overall]
</doc_summary>
\end{lstlisting}

For documents that exceed a single context window, \modelname{} first applies a local structure prompt to identify topic boundaries within each page segment:

\begin{lstlisting}[language={}, caption={Prompt excerpt for local structure extraction.}]
You are analyzing a segment of a larger document.
You will be provided with summaries for pages {START} to {END}.

Your task is to identify the major thematic topics covered in this page range.

Guidelines:
1. Identify distinct thematic transitions within this segment.
2. Group consecutive pages that cover the same theme into a single section entry.
3. Use concise, descriptive topic labels.
4. Do not speculate about content outside this page range.

Output:
<local_structure>
Pages X-Y: [section topic]
Pages Y-Z: [section topic]
...
</local_structure>
\end{lstlisting}

The merge prompt then synthesizes multiple batch-level structures into unified \texttt{doc\_type}, \texttt{doc\_structure}, and \texttt{doc\_summary}, while merging adjacent topics, normalizing section names, and preserving the original page ranges.

\noindent\textbf{Page Feedback for Page Belief Memory}
The reasoning prompt in \modelname{} not only generates the final decision, but also explicitly assesses whether each retrieved page is useful for the current question. 
Specifically, the model is required to output \texttt{useful\_pages} and \texttt{irrelevant\_pages}: the former indicates pages that contain valid evidence or provide useful reasoning cues, while the latter indicates retrieved pages that do not contribute useful information. 
These two fields serve as positive and negative feedback signals for subsequent Bayesian Page Belief updating.

\begin{lstlisting}[language={}, caption={Prompt excerpt for page-level feedback.}]
Based on your analysis, provide the following page assessment
before your final answer:

<useful_pages>
[Comma-separated page numbers that contained relevant information,
e.g. 3,7. Leave empty if none.]
</useful_pages>

<irrelevant_pages>
[Comma-separated page numbers that were not helpful,
e.g. 5,12. Leave empty if none.]
</irrelevant_pages>
\end{lstlisting}

This design converts the reasoner's intermediate judgments into explicit page-level supervision signals. 
Pages listed in \texttt{useful\_pages} receive increased relevance confidence, whereas pages listed in \texttt{irrelevant\_pages} are assigned lower priority in later retrieval rounds. 
In this way, reasoning feedback from each round is written back into \textit{Page Belief Memory} and used to guide subsequent candidate-page selection.

\noindent\textbf{Query Update and Question Episodic Memory}
When the current pages are insufficient to answer the question but the model infers that the answer may appear in unretrieved parts of the document, the QA prompt asks for both \texttt{query\_update} and \texttt{notes}. 
The \texttt{query\_update} is a refined retrieval query that preserves the original intent of the question, while \texttt{notes} summarize the evidence observed so far, the information still missing, and the rationale for issuing a new query. 
These notes are written into \textit{Question Episodic Memory} to maintain the cross-round reasoning trajectory.

\begin{lstlisting}[language={}, caption={Prompt excerpt for query update and episodic notes.}]
If based on the document summary, you believe the answer exists in
other parts of the document that haven't been retrieved yet:

<query_update>
[Provide a rewritten long query that PRESERVES THE ORIGINAL MEANING
of the question but adds specific details or keywords to help retrieve
new relevant pages.]
</query_update>

<notes>
[Provide concise notes about what you've learned so far, what information
is still missing, and your reasoning for the updated query.]
</notes>
\end{lstlisting}

The prompt further instructs the model to avoid repeatedly requesting pages that have already been retrieved or visual elements that are already visible, thereby reducing ineffective retrieval loops.

\begin{lstlisting}[language={}, caption={Prompt excerpt for avoiding repeated retrieval.}]
CRITICAL INSTRUCTION: First carefully check if:
- The pages listed in <retrieved_pages> are already the specific pages
  that would contain the answer.
- The specific tables, figures, charts, or other elements referenced
  in the question are already visible in the current images.

Do not request these same pages or elements again in a query update.
\end{lstlisting}

This constraint allows \textit{Question Episodic Memory} to record not only what should be searched next, but also what has already been ruled out and what remains missing. 
As a result, subsequent retrieval rounds can continue narrowing the search space along the existing evidence chain rather than repeating stateless retrieval.

\begin{table}[t]
\centering
\small
\begin{tabular}{lcc}
\toprule
\textbf{Iteration} & \textbf{Round Acc.} & \textbf{Cumulative Acc.} \\
\midrule
Iter 1 & 72.0 & 60.6 \\
Iter 2 & 59.1 & 65.5 \\
Iter 3 & 46.2 & 69.0 \\
\bottomrule
\end{tabular}
\caption{Accuracy improvement of \modelname{} across retrieval--reasoning iterations on MMLongBench-Doc.}
\label{tab:iter_acc}
\end{table}

\noindent\textbf{Adaptive-Granularity Evidence Access}
To support fine-grained reading over information-dense regions such as tables and charts, the QA prompt in \modelname{} allows high-resolution local crops to be appended after full-page images. 
The model is instructed to jointly analyze rendered page images, extracted text, and possible table crops, thereby preserving page-level context while recovering local visual details.

\begin{lstlisting}[language={}, caption={Prompt excerpt for adaptive-granularity evidence access.}]
Images of the retrieved pages are attached at the end of this prompt.
The raw text extracted from these images is provided in the <page_text> tag above.

You must analyze BOTH the visual images AND the extracted text,
along with the <document_summary>, to fully understand the document
and answer the question accurately.

For pages containing tables, additional high-resolution table crop images
may be appended after the page images -- use these for precise table value reading.
\end{lstlisting}

This design corresponds to adaptive-granularity evidence access: \modelname{} first localizes evidence at the page level, and then supplements region-level inputs when a page contains locally dense visual evidence, mitigating information loss caused by full-page downscaling.

\begin{table*}[t!]
\centering
\resizebox{\textwidth}{!}{%
\begin{tabular}{l r r r r r r}
\toprule
\textbf{Stage} & \textbf{Calls} & \textbf{Avg. Time/Doc} & \textbf{Total Time} & \textbf{Prompt Tokens (avg/call)} & \textbf{Completion Tokens (avg/call)} & \textbf{Storage} \\
\midrule
Visual Embedding (ColQwen2.5) & 10 & 6.7 s & 0.02 h & -- & -- & 1.2 GB \\
Page Summary Generation & 272 & 0.05 h & 0.47 h & 2326 & 318 & 6.5 MB \\
Doc Schema Memory (batch + merge) & 24 & 8.6 s & 0.02 h & 3275 & 202 & 191.3 kB \\
Total & -- & -- & 0.51 h & 711373 & 91412 & 1.2 GB \\
\bottomrule
\end{tabular}%
}
\caption{Computational cost of each stage.}
\label{tab:computational_cost}
\end{table*}

\begin{table*}[t!]
\centering
\resizebox{\textwidth}{!}{%
\begin{tabular}{l r r r r r r r}
\toprule
\textbf{Metric} & \textbf{API Calls} & \textbf{Avg. Calls / Question} & \textbf{Avg. Images / Call} & \textbf{Avg. Input Tokens / Call} & \textbf{Avg. Output Tokens / Call} & \textbf{Avg. Latency / Call (s)} & \textbf{Peak VRAM (GB)} \\
\midrule
Reasoning VLM & 87 & 1.21 & 5.2 & 11541 & 1121 & 20.41 & 33.95 \\
\bottomrule
\end{tabular}%
}
\caption{Reasoning VLM performance metrics.}
\label{tab:reasoning_vlm_metrics}
\end{table*}

\begin{table}[t]
\centering
\small
\resizebox{\columnwidth}{!}{%
\begin{tabular}{lcc}
\toprule
\textbf{Metric} & \textbf{SimpleDoc} & \textbf{\modelname{}} \\
\midrule
Iterations per question & 3.00 (fixed) & 1.24 (avg.) \\
Relative computation cost & 1.00$\times$ & 0.41$\times$ \\
Relative efficiency & 1.00$\times$ & 2.40$\times$ \\
MMLongBench-Doc accuracy & 60.6 & 71.3 \\
\bottomrule
\end{tabular}%
}
\caption{Iteration efficiency comparison on MMLongBench-Doc. Relative computation cost is estimated by LLM/VLM calls induced by retrieval--reasoning iterations, and relative efficiency is its inverse.}
\label{tab:iter_efficiency}
\end{table}

\begin{table}[t]
\centering
\small
\resizebox{\columnwidth}{!}{%
\begin{tabular}{lccccc c}
\toprule
\multirow{2}{*}{\textbf{Method}} 
& \multicolumn{5}{c}{\textbf{LongURL}} 
& \textbf{PaperTab} \\
\cmidrule(lr){2-6} \cmidrule(lr){7-7}
& \textbf{Text} & \textbf{Layout} & \textbf{Table} & \textbf{Figure} & \textbf{ALL} & \textbf{ALL} \\
\midrule
w/o Bayesian 
& 84.8 & 73.1 & 81.0 & 77.7 & \(79.2 \pm 0.2\) & \(78.6 \pm 0.2\) \\
w/o All Memory 
& 85.3 & 74.0 & 81.2 & 77.7 & \(79.4 \pm 0.2\) & \(78.4 \pm 0.3\) \\
w/o Adaptive-granularity 
& 86.5 & 75.7 & 82.6 & 81.3 & \(80.8 \pm 0.3\) & \(79.2 \pm 0.2\) \\
\midrule
\textbf{Ours (Full)} 
& \textbf{86.1} & \textbf{74.4} & \textbf{83.9} & \textbf{81.6} 
& \(\mathbf{81.2 \pm 0.3}\) & \(\mathbf{80.4 \pm 0.2}\) \\
\bottomrule
\end{tabular}%
}
\caption{
Additional ablation results on LongURL and PaperTab. 
}
\label{tab:additional_ablation}
\end{table}

\section{Additional Experimental Results}
\label{app:additional_results}

\subsection{Accuracy Gains across Iterations}

Table~\ref{tab:iter_acc} reports how the cumulative accuracy of \modelname{} changes across retrieval--reasoning iterations on MMLongBench-Doc. The first iteration already resolves most relatively direct questions, reaching a cumulative accuracy of 60.6\%. The second iteration further improves the accuracy to 65.5\%, suggesting that query refinement and page belief updating can recover evidence missed by the initial retrieval. The third iteration still brings additional gains, but with a smaller margin, indicating that most recoverable evidence is located within the first two rounds and later iterations mainly provide marginal refinement. Overall, these results show that \modelname{} benefits from effective evidence recovery within a small number of iterations, rather than simply relying on more retrieval rounds.

\subsection{Iteration Efficiency}

Table~\ref{tab:iter_efficiency} compares the iteration efficiency of \modelname{} and SimpleDoc on MMLongBench-Doc. 
SimpleDoc performs a fixed three-round retrieval--reasoning process for every question, whereas \modelname{} continues retrieval only when the current evidence is insufficient, resulting in an average of 1.24 rounds per question. We compute relative computation cost by the number of LLM/VLM calls incurred in retrieval--reasoning loops; under this metric, \modelname{} requires only 0.41$\times$ the cost of SimpleDoc, corresponding to about 2.4$\times$ relative efficiency. This suggests that the gains of \modelname{} do not come from increasing the number of iterations, but from on-demand iteration, structured memory, and page belief updating, which enable more effective evidence exploration at lower invocation cost.

\subsection{Additional Ablations}

Table~\ref{tab:additional_ablation} further reports supplementary ablation results on LongURL and PaperTab, aiming to verify the robustness of the core components across different document scenarios. Removing either Bayesian page belief updating or the full memory module leads to consistent performance degradation: the overall accuracy drops by 2.0 and 1.8 points on LongURL, and by 1.8 and 2.0 points on PaperTab, respectively. This indicates that cross-round page belief updating and structured memory are not only effective on a single benchmark, but also provide stable gains for long-document page selection and evidence integration. A closer look at the fine-grained categories on LongURL shows that both ablations lead to more evident drops on Table and Figure subsets, suggesting that dynamic page states and memory context are especially important for locating cross-page visual evidence.

On the other hand, removing adaptive-granularity evidence access causes only a small drop in the overall LongURL score, but leads to more noticeable degradation on the Table subset and PaperTab, with drops of 1.3 and 1.2 points, respectively. Since these settings rely more heavily on table cells, local numerical values, and fine-grained visual structures, full-page inputs alone may compress or miss key details. These results further suggest that the main gains of \modelname{} come from structured memory and dynamic page belief updating, while adaptive-granularity evidence access mainly strengthens its ability to handle table-intensive and local-detail-dependent questions.

\begin{table}[t]
\centering
\resizebox{\columnwidth}{!}{%
\begin{tabular}{l c c c c c c c}
\toprule
\textbf{Method} & \textbf{TXT} & \textbf{LAY} & \textbf{CHA} & \textbf{TAB} & \textbf{FIG} & \textbf{UNA} & \textbf{ALL} \\
\midrule
Ours (Full, multi-modal) & 70.5 & 61.3 & 67.8 & 73.3 & 63.8 & 78.8 & 71.3 \\
OCR-only (w/o visual modality) & 65.1 & 37.5 & 34.9 & 47.8 & 40.9 & 72.6 & 58.6 \\
Vision-only (w/o text modality) & 67.7 & 54.8 & 66.5 & 72.2 & 60.5 & 79.5 & 69.3 \\
w/o fine-grained visual crops & 68.2 & 58.9 & 64.2 & 68.7 & 63.2 & 78.8 & 69.9 \\
\bottomrule
\end{tabular}%
}
\caption{Ablation study on modality components.}
\label{tab:ablation_modality}
\end{table}

\begin{table}[t]
\centering
\resizebox{\columnwidth}{!}{%
\begin{tabular}{l c c c c c c c}
\toprule
\textbf{Method} & \textbf{TXT} & \textbf{LAY} & \textbf{CHA} & \textbf{TAB} & \textbf{FIG} & \textbf{UNA} & \textbf{ALL} \\
\midrule
Thompson Sampling & 70.53 & 61.34 & 67.80 & 73.27 & 63.77 & 78.79 & 71.28 \\
Greedy Selection & 67.43 & 53.78 & 61.93 & 70.18 & 58.22 & 76.96 & 68.62 \\
\bottomrule
\end{tabular}%
}
\caption{Comparison of selection strategies.}
\label{tab:selection_strategies}
\end{table}

\subsection{Descriptive Statistics}

For the additional ablations in Table~\ref{tab:additional_ablation}, we report mean accuracy from repeated evaluations, with uncertainty shown as \(\pm\) values. The full \modelname{} achieves \(81.2 \pm 0.3\) on LongURL and \(80.4 \pm 0.2\) on PaperTab. Removing Bayesian page belief updating decreases the scores to \(79.2 \pm 0.2\) and \(78.6 \pm 0.2\), while removing all memory components yields \(79.4 \pm 0.2\) and \(78.4 \pm 0.3\), respectively. Without adaptive-granularity evidence access, the scores are \(80.8 \pm 0.3\) on LongURL and \(79.2 \pm 0.2\) on PaperTab, showing a smaller but still consistent degradation. The small variances across repeated evaluations indicate stable results under the same evaluation protocol, and the consistent gaps between the full system and ablated variants suggest that the improvements are not driven by outlier runs or evaluation noise.

\subsection{Computation Cost Analysis}

To further evaluate the deployment cost, we conducted experiments on both offline preprocessing cost and online inference cost. We report the computational cost of the offline preprocessing stage for 10 documents.

As shown in Table~\ref{tab:computational_cost} and Table~\ref{tab:reasoning_vlm_metrics}, the construction of the tri-level memory takes 0.51 hours in total, with the major cost coming from page summary generation (0.47 hours), while visual embedding extraction and Document Schema Memory construction require only 0.02 hours each. The storage cost is dominated by visual embeddings (1.2 GB), whereas the generated textual memory remains lightweight.

For online inference, the model requires only 1.21 VLM calls per question on average. Each call processes an average of 5.2 images, 11.5K input tokens, and 1.1K output tokens, with an average latency of 20.41 seconds. These results indicate that the additional overhead introduced by memory construction and evidence discovery mechanisms is controllable. Moreover, since offline-constructed memory can be reused across multiple queries, its cost can be amortized over many question-answering sessions.

\subsection{Human Verification of Automatic Evaluation Results}
In this work, we follow recent baselines by adopting GPT-4.1 as the evaluator. However, we additionally conduct an analysis on the agreement between GPT-4.1 and human judges. Specifically, we randomly sampled 300 questions from the evaluation results of MMLongBench. We manually reviewed each predicted answer to determine whether it exactly matched the ground-truth answer, and compared the human judgments with the binary decisions produced by GPT-4.1. The overall agreement between the two is 96.7\%, with a Cohen's kappa coefficient of 0.92, indicating an almost perfect level of consistency.

\subsection{Modal Ablation Study}
Due to the heterogeneity of evidence in long documents, while precise numerical values and long-text semantics still rely on the text modality, visually intensive information such as tables and figures depends on visual structures that are difficult to comprehend from flattened text alone. Therefore, the multimodal mechanism plays an important role in our framework.

Here, we present the modal ablation study on the MMLongBench benchmark to demonstrate its contribution. As shown in Table~\ref{tab:ablation_modality}, removing either modality leads to an overall performance drop, with the most significant degradation observed in question types that heavily depend on the corresponding modality. This indicates that the multimodal mechanism indeed helps improve performance. Moreover, removing fine-grained visual cropping leads to a noticeable performance decline in table-related questions, which validates the complementary value of local visual evidence to page-level representations.

\subsection{Necessity Analysis of Thompson Sampling}
Thompson Sampling leverages posterior uncertainty to preserve exploration opportunities for pages that have not yet been sufficiently verified, making it particularly suitable for multi-round retrieval scenarios with iterative feedback. In contrast, Greedy Selection ranks pages solely according to their posterior means, making it prone to prematurely fixing early selections and overlooking potentially relevant pages.

As shown in Table~\ref{tab:selection_strategies}, we replaced Thompson Sampling with Greedy Selection, which always ranks pages according to the posterior mean. The experimental results show that removing Thompson Sampling decreases the overall performance from 71.28 to 68.62, indicating that the exploration-exploitation balance it provides is necessary for multi-round evidence discovery.


\end{document}